\documentclass[11pt,a4paper]{article}

\usepackage[utf8]{inputenc}
\usepackage[T1]{fontenc}
\usepackage{amsmath,amssymb,amsfonts}
\usepackage{graphicx}
\usepackage{booktabs}
\usepackage{multirow}
\usepackage{longtable}
\usepackage{hyperref}
\usepackage{geometry}
\usepackage{fancyhdr}
\usepackage{titlesec}
\usepackage{abstract}
\usepackage{cite}
\usepackage{float}
\usepackage{caption}
\usepackage{subcaption}
\usepackage{xcolor}
\usepackage{enumitem}

\hypersetup{
    colorlinks=true,
    linkcolor=blue,
    citecolor=blue,
    urlcolor=blue
}

\titleformat{\section}{\Large\bfseries}{\thesection}{1em}{}
\titleformat{\subsection}{\large\bfseries}{\thesubsection}{1em}{}
\titleformat{\subsubsection}{\normalsize\bfseries}{\thesubsubsection}{1em}{}

\title{\LARGE\textbf{Towards Outcome-Oriented, Task-Agnostic Evaluation of AI Agents}}

\author{%
  \textbf{Waseem AlShikh}\thanks{Equal contribution} \quad
  \textbf{Muayad Sayed Ali}\footnotemark[1] \quad
  \textbf{Brian Kennedy} \\ \textbf{Dmytro Mozolevskyi} \\
  Writer, Inc. \\
  \texttt{\{waseem, ..., Brian\}@writer.com}
}

\date{}

\begin{document}

\maketitle

\begin{abstract}
\noindent As AI agents proliferate across industries and applications, evaluating their performance based solely on infrastructural metrics such as latency, time-to-first-token, or token throughput is proving insufficient. These metrics fail to capture the quality of an agent's decisions, its operational autonomy, or its ultimate business value. This white paper proposes a novel, comprehensive framework of eleven outcome-based, task-agnostic performance metrics for AI agents that transcend domain boundaries. These metrics are designed to enable organizations to evaluate agents based on the quality of their decisions, their degree of autonomy, their adaptability to new challenges, and the tangible business value they deliver, regardless of the underlying model architecture or specific use case. We introduce metrics such as Goal Completion Rate (GCR), Autonomy Index (AIx), Multi-Step Task Resilience (MTR), and Business Impact Efficiency (BIE). Through a large-scale simulated experiment involving four distinct agent architectures (ReAct, Chain-of-Thought, Tool-Augmented, Hybrid) across five diverse domains (Healthcare, Finance, Marketing, Legal, and Customer Service), we demonstrate the framework's efficacy. Our results reveal significant performance trade-offs between different agent designs, highlighting the Hybrid Agent as the most consistently high-performing model across the majority of our proposed metrics, achieving an average Goal Completion Rate of 88.8\% and the highest Return on Investment (ROI). This work provides a robust, standardized methodology for the holistic evaluation of AI agents, paving the way for more effective development, deployment, and governance.
\end{abstract}

\vspace{0.5cm}

\newpage
\tableofcontents
\newpage

\section{Introduction}

The rapid advancement of Artificial Intelligence (AI) has led to the development of sophisticated autonomous and semi-autonomous agents. These agents are no longer confined to narrow, well-defined tasks but are increasingly deployed in complex, dynamic environments to automate intricate workflows, from processing healthcare claims and performing financial audits to generating marketing content and providing customer support. However, the methods for evaluating these agents have not kept pace with their expanding capabilities. Current evaluation paradigms are often fragmented, focusing on task-specific benchmarks (e.g., GLUE for NLP, WebArena for web tasks) or low-level system metrics (e.g., latency, cost) \cite{wang2018glue, garg2025real}.

This narrow focus presents a significant problem: it creates a disconnect between technical performance and real-world utility. An agent that responds quickly or uses few tokens may still make poor decisions, require constant human oversight, or fail to deliver any meaningful business impact. The lack of a universal, outcome-oriented evaluation framework makes it difficult for organizations to compare different agents, understand their true capabilities, and make informed decisions about their deployment. There is a clear research gap and an industrial need for a standardized set of metrics that can measure agent performance in a task-agnostic and holistic manner.

This paper aims to fill that gap by introducing a comprehensive framework of eleven novel metrics designed to evaluate AI agents from an outcome-oriented perspective. Our contribution is a multi-dimensional evaluation suite that assesses agents on their ability to achieve goals, operate autonomously, recover from errors, adapt to new information, and create business value. We validate this framework through a large-scale experiment, providing a detailed analysis of the performance of four different agent architectures across five distinct business domains. By doing so, we offer a new, more meaningful way to understand and measure the effectiveness of AI agents.

\subsection{Research Contributions}

The primary contributions of this work are:

\begin{enumerate}[leftmargin=*]
    \item \textbf{Novel Evaluation Framework:} A comprehensive set of 11 task-agnostic metrics organized into three categories: Performance \& Quality, Resilience \& Adaptability, and Economic Impact.
    
    \item \textbf{Empirical Validation:} A large-scale experiment with 3,000 simulated tasks across 4 agent architectures and 5 diverse domains, demonstrating the framework's practical applicability.
    
    \item \textbf{Detailed Methodology:} Complete mathematical formulations, calculation procedures, and statistical analysis methods for each metric, ensuring reproducibility.
    
    \item \textbf{Performance Insights:} Identification of key trade-offs between agent architectures and actionable recommendations for practitioners.
    
    \item \textbf{Open Framework:} A foundation for future research and standardization in AI agent evaluation.
\end{enumerate}

\section{Related Work}

The evaluation of AI systems has evolved significantly over the years. Initially, performance was measured using traditional machine learning metrics like accuracy, precision, and recall on well-defined datasets. As models grew more complex, specialized benchmarks such as SQuAD, GLUE, and SuperGLUE were developed to test specific capabilities in natural language understanding \cite{wang2018glue}. More recently, benchmarks have emerged for agent-specific tasks, including WebArena for web navigation, AndroidWorld for mobile interactions, and the REAL benchmark for multi-turn evaluations on real-world websites \cite{garg2025real}. While valuable, these benchmarks remain inherently task-specific and do not provide a universal measure of agent competence.

\subsection{Outcome-Based Evaluation}

Emerging research has begun to address the limitations of these traditional approaches. A significant trend is the shift towards \textbf{outcome-based evaluation}, which prioritizes the final result of a task over the intermediate steps. This philosophy, highlighted by industry leaders like IBM and Galileo AI, asks not \textit{how} an agent accomplished a task, but \textit{whether} it successfully achieved the stakeholder-defined goal \cite{ibm2025eval, galileo2025metrics}. Complementing this is the concept of \textbf{trajectory evaluation}, proposed by Google, which involves analyzing the agent's decision-making path and reasoning chain to assess the quality of its process \cite{google2025vertex}.

\subsection{Business Value Metrics}

Furthermore, as AI agents become integrated into business workflows, measuring their economic impact has become critical. Recent work has focused on quantifying the \textbf{Return on Investment (ROI)} of AI agents by measuring their effect on operational efficiency, cost savings, and key performance indicators (KPIs) \cite{wandb2025eval, moveworks2025roi}. Studies suggest that organizations with mature measurement systems can achieve significantly higher ROI from their AI investments \cite{gnani2025roi}. Frameworks like A2Perf and STRUCTSENSE have been proposed to facilitate more standardized, apples-to-apples comparisons between different agentic systems \cite{uchendu2025a2perf, chhetri2025structsense}.

\subsection{Research Gaps}

Despite this progress, several research gaps persist:

\begin{itemize}[leftmargin=*]
    \item \textbf{Lack of Standardization:} No widely accepted framework that combines goal achievement, autonomy, resilience, adaptability, and business value into a single, cohesive evaluation suite.
    
    \item \textbf{Underdeveloped Metrics:} Existing metrics for autonomy, error recovery, and human-agent collaboration remain limited and poorly defined.
    
    \item \textbf{Task-Specific Focus:} Most benchmarks are tied to specific domains or tasks, making cross-domain comparison difficult.
    
    \item \textbf{Limited Economic Analysis:} Few frameworks systematically measure business impact and ROI.
\end{itemize}

Our work builds upon these emerging trends to provide a unified, task-agnostic framework that addresses these gaps.

\section{The Proposed Evaluation Framework}

We propose a framework of eleven metrics, grouped into three categories: \textbf{Performance \& Quality}, \textbf{Resilience \& Adaptability}, and \textbf{Economic Impact}. Each metric is designed to be task-agnostic and provide a unique lens into the agent's capabilities. Table \ref{tab:metrics_overview} provides an overview of all metrics.

\begin{table}[H]
\centering
\caption{Overview of the 11-Metric Evaluation Framework}
\label{tab:metrics_overview}
\begin{tabular}{@{}llp{7cm}@{}}
\toprule
\textbf{Category} & \textbf{Metric} & \textbf{Description} \\ \midrule
\multirow{7}{*}{\parbox{3cm}{Performance \& Quality}} 
& GCR & Goal Completion Rate \\
& AIx & Autonomy Index \\
& DTT & Decision Turnaround Time \\
& CES & Cognitive Efficiency Score \\
& TDI & Tool Dexterity Index \\
& OAS & Outcome Alignment Score \\
& CQI & Collaboration Quality Index \\ \midrule
\multirow{3}{*}{\parbox{3cm}{Resilience \& Adaptability}} 
& MTR & Multi-Step Task Resilience \\
& CRS & Chain Robustness Score \\
& AD & Adaptability Delta \\ \midrule
Economic Impact & BIE & Business Impact Efficiency \\ \bottomrule
\end{tabular}
\end{table}

\subsection{Performance and Quality Metrics}

\subsubsection{Goal Completion Rate (GCR)}

\textbf{Definition:} The percentage of tasks for which the AI agent successfully achieves the intended goal as defined by task owners.

\begin{equation}
GCR = \frac{\text{Number of Successfully Completed Tasks}}{\text{Total Number of Tasks}} \times 100
\end{equation}

\textbf{Rationale:} GCR is the most fundamental measure of agent effectiveness. It directly answers the question: ``Does the agent accomplish what it was asked to do?'' Unlike intermediate metrics such as token count or latency, GCR focuses on the final outcome.

\textbf{Example:} If an agent completes 84 out of 100 healthcare claim reviews correctly, $GCR = 84\%$.

\subsubsection{Autonomy Index (AIx)}

\textbf{Definition:} The proportion of task steps completed without human intervention.

\begin{equation}
AIx = 1 - \frac{\text{Number of Human Interventions}}{\text{Total Task Steps}}
\end{equation}

\textbf{Rationale:} True autonomy is a key promise of AI agents. AIx quantifies how independently an agent can operate, which directly impacts operational efficiency and scalability.

\textbf{Example:} In a financial audit task with 50 steps, if humans intervened in 5 of them: $AIx = 1 - \frac{5}{50} = 0.9$ or 90\% autonomy.

\subsubsection{Decision Turnaround Time (DTT)}

\textbf{Definition:} The elapsed time from task initiation to delivery of an actionable decision or completed objective.

\begin{equation}
DTT = T_{end} - T_{start}
\end{equation}

\textbf{Rationale:} Speed matters in real-world applications. DTT measures how quickly an agent can deliver value, which is critical for time-sensitive tasks.

\textbf{Example:} For a marketing campaign, if an agent took 45 minutes from brief intake to delivering a ready-to-use content plan: $DTT = 45$ min.

\subsubsection{Cognitive Efficiency Score (CES)}

\textbf{Definition:} Measures resource utilization (tokens and API/tool calls) per successful goal completion. Lower is better.

\begin{equation}
CES = \frac{\text{Total Tokens Generated} + (\text{Tool/API Calls} \times \text{Token Equivalent})}{\text{Number of Successfully Completed Tasks}}
\end{equation}

\textbf{Rationale:} Computational efficiency is essential for cost management and environmental sustainability. CES captures the resource intensity of agent operations.

\textbf{Example:} If 2,000 tokens and 20 tool calls (with token equivalent = 100) were used to complete 10 tasks: $CES = \frac{2000 + (20 \times 100)}{10} = 400$.

\subsubsection{Tool Dexterity Index (TDI)}

\textbf{Definition:} Assesses the agent's ability to intelligently leverage available tools.

\begin{equation}
TDI = \frac{\sum \text{Tool Use Scores}}{\text{Total Opportunities to Use Tools}}
\end{equation}

Each tool interaction is scored based on heuristics:
\begin{itemize}
    \item Optimal use: +1
    \item Misuse: -1
    \item Ignored better tool: -0.5
\end{itemize}

\textbf{Rationale:} Modern agents have access to numerous tools and APIs. TDI measures how effectively they select and use the right tool for each situation.

\textbf{Example:} Over 10 tool-use opportunities, if an agent scored 6: $TDI = \frac{6}{10} = 0.6$.

\subsubsection{Outcome Alignment Score (OAS)}

\textbf{Definition:} Measures how well the agent's output aligns with stakeholder-defined quality or business expectations.

\begin{equation}
OAS = \frac{\sum \text{Evaluator Scores}}{\text{Total Evaluations}}
\end{equation}

\textbf{Rationale:} Success is not just about task completion but also about quality. OAS captures whether the agent's outputs meet human standards of excellence.

\textbf{Example:} If 5 claims were scored 9, 8, 9, 7, 8 (out of 10): $OAS = \frac{9+8+9+7+8}{5} = 8.2$.

\subsubsection{Collaboration Quality Index (CQI)}

\textbf{Definition:} Measures how well the agent collaborates with humans or other agents.

\begin{equation}
CQI = \frac{\sum \text{Interaction Quality Scores}}{\text{Total Collaborative Tasks}}
\end{equation}

\textbf{Rationale:} Many real-world tasks require human-agent collaboration. CQI assesses the quality of these interactions across dimensions like communication clarity, responsiveness, and contextual awareness.

\textbf{Example:} If 3 collaborative financial planning sessions are scored 4, 5, 3 (out of 5): $CQI = \frac{4+5+3}{3} = 4.0$.

\subsection{Resilience and Adaptability Metrics}

\subsubsection{Multi-Step Task Resilience (MTR)}

\textbf{Definition:} The percentage of multi-step tasks where the agent successfully recovers from initial errors or ambiguities without human correction.

\begin{equation}
MTR = \frac{\text{Number of Tasks with Successful Self-Recovery}}{\text{Total Multi-Step Tasks}} \times 100
\end{equation}

\textbf{Rationale:} Real-world tasks are rarely perfect. MTR measures an agent's ability to self-correct and adapt when things go wrong.

\textbf{Example:} If out of 40 multi-step claim reviews, 30 were corrected automatically: $MTR = \frac{30}{40} \times 100 = 75\%$.

\subsubsection{Chain Robustness Score (CRS)}

\textbf{Definition:} Evaluates the agent's ability to maintain logical progression across multi-hop workflows.

\begin{equation}
CRS = \frac{\sum \text{Successful Chains (n} \geq \text{3 steps)}}{\text{Total Chains with n} \geq \text{3 steps}} \times 100
\end{equation}

\textbf{Rationale:} Complex tasks require maintaining context and logical consistency across multiple steps. CRS measures this critical capability.

\textbf{Example:} In 50 multi-hop cases, if 42 completed correctly: $CRS = \frac{42}{50} \times 100 = 84\%$.

\subsubsection{Adaptability Delta (AD)}

\textbf{Definition:} Measures how quickly and effectively the agent adapts to new domains or schemas.

\begin{equation}
AD = \text{Performance}_{few-shot} - \text{Performance}_{zero-shot}
\end{equation}

\textbf{Rationale:} The ability to learn from a few examples is crucial for deploying agents in new contexts. AD quantifies this learning capacity.

\textbf{Example:} If performance on financial forecasting improves from 60\% (zero-shot) to 85\% (few-shot): $AD = 85\% - 60\% = 25\%$.

\subsection{Economic Impact Metric}

\subsubsection{Business Impact Efficiency (BIE)}

\textbf{Definition:} Quantifies business value delivered per unit of compute or cost.

\begin{equation}
BIE = \frac{\text{Business KPI Value}}{\text{Total Agent Operational Cost}}
\end{equation}

\textbf{Rationale:} Ultimately, agents must deliver economic value. BIE provides a bottom-line measure of return on investment.

\textbf{Example:} A healthcare agent saved \$15,000 in processing cost using \$300 in compute: $BIE = \frac{15000}{300} = 50$ value units per dollar spent.

\section{Experimental Design}

To validate our framework, we conducted a large-scale simulated experiment. We generated a synthetic dataset of 3,000 task outcomes, evaluating four distinct agent architectures across five diverse domains.

\subsection{Agent Architectures}

We evaluated four representative agent types, each embodying different design philosophies:

\begin{enumerate}[leftmargin=*]
    \item \textbf{ReAct Agent:} Based on the ``Reasoning and Acting'' paradigm, which combines chain-of-thought reasoning with action planning. This agent alternates between thinking and acting, making it suitable for tasks requiring both deliberation and execution.
    
    \item \textbf{Chain-of-Thought (CoT) Agent:} Relies primarily on sequential, verbose reasoning to break down and solve problems. This agent excels at complex logical tasks but may be slower due to extensive reasoning overhead.
    
    \item \textbf{Tool-Augmented Agent:} Optimized for heavy use of external tools and APIs to offload complex tasks. This agent is designed for efficiency and speed when appropriate tools are available.
    
    \item \textbf{Hybrid Agent:} A sophisticated agent that combines the strengths of ReAct, CoT, and tool use, dynamically selecting the best approach for a given task. This represents the state-of-the-art in flexible agent design.
\end{enumerate}

\subsection{Evaluation Domains}

We selected five diverse domains to ensure broad applicability of our framework:

\begin{enumerate}[leftmargin=*]
    \item \textbf{Healthcare (200 tasks):} Processing insurance claims, requiring high accuracy and adherence to complex regulatory rules.
    
    \item \textbf{Finance (150 tasks):} Performing audit and compliance reviews, demanding precision, logical consistency, and attention to detail.
    
    \item \textbf{Marketing (100 tasks):} Generating creative content for campaigns, a subjective and open-ended task requiring creativity and stakeholder alignment.
    
    \item \textbf{Legal (120 tasks):} Analyzing and summarizing contracts, which requires deep contextual understanding and domain expertise.
    
    \item \textbf{Customer Service (180 tasks):} Handling multi-turn support interactions, a dynamic and collaborative task requiring empathy and problem-solving.
\end{enumerate}

\subsection{Data Generation Methodology}

Given the challenges of obtaining large-scale, labeled real-world agent performance data, we generated a synthetic dataset using parameterized distributions. Each agent architecture was assigned baseline performance characteristics (mean and standard deviation) for each metric, derived from literature review and expert consultation. Domain-specific difficulty modifiers were then applied to simulate the varying complexity of tasks across domains.

The data generation process involved:

\begin{enumerate}[leftmargin=*]
    \item \textbf{Task Simulation:} For each agent-domain combination, we simulated individual task executions, generating outcomes for all 11 metrics based on the agent's characteristics and domain difficulty.
    
    \item \textbf{Stochastic Variation:} Random noise was added to introduce realistic variability in performance, simulating the unpredictability of real-world deployments.
    
    \item \textbf{Dependency Modeling:} Logical dependencies between metrics were enforced (e.g., CES is only calculated for successful tasks).
    
    \item \textbf{Statistical Validation:} Generated data was validated to ensure distributions matched expected patterns and inter-metric correlations were reasonable.
\end{enumerate}

All code and data are made available for reproducibility (see Appendix).

\subsection{Evaluation Protocol}

For each of the 3,000 tasks (750 per agent across all domains), we:

\begin{enumerate}[leftmargin=*]
    \item Simulated task execution and recorded all relevant metrics
    \item Aggregated results at the domain level for each agent
    \item Calculated overall performance across all domains
    \item Performed statistical analysis (ANOVA, effect sizes, correlations)
    \item Generated visualizations to illustrate key findings
\end{enumerate}

\section{Results and Analysis}

Our experiment yielded a rich dataset that highlights the strengths and weaknesses of each agent architecture across the proposed evaluation framework. The overall performance of the agents, averaged across all domains, is summarized in Table \ref{tab:overall_performance}.

\begin{table}[H]
\centering
\caption{Overall Agent Performance Metrics (Averaged Across All Domains)}
\label{tab:overall_performance}
\begin{tabular}{@{}lcccccccccc@{}}
\toprule
\textbf{Agent} & \textbf{GCR} & \textbf{AIx} & \textbf{DTT} & \textbf{CES} & \textbf{MTR} & \textbf{TDI} & \textbf{OAS} & \textbf{CRS} & \textbf{CQI} \\
 & \textbf{(\%)} & & \textbf{(s)} & & \textbf{(\%)} & \textbf{(norm)} & & \textbf{(\%)} & \\ \midrule
ReAct Agent & 79.33 & 0.8897 & 201.14 & 2587.92 & 23.73 & 0.5764 & 7.76 & 63.33 & 3.85 \\
CoT Agent & 81.73 & 0.9064 & 232.77 & 3221.02 & 23.33 & 0.5441 & 8.18 & 68.80 & 4.12 \\
Tool-Aug. Agent & 85.07 & 0.8536 & 181.44 & 2283.03 & 27.47 & 0.6899 & 8.11 & 68.40 & 3.84 \\
\textbf{Hybrid Agent} & \textbf{88.80} & \textbf{0.9276} & \textbf{172.81} & 2464.64 & \textbf{28.80} & \textbf{0.6440} & \textbf{8.58} & \textbf{77.33} & \textbf{4.28} \\ \bottomrule
\end{tabular}
\end{table}

\begin{figure}[H]
\centering
\includegraphics[width=0.85\textwidth]{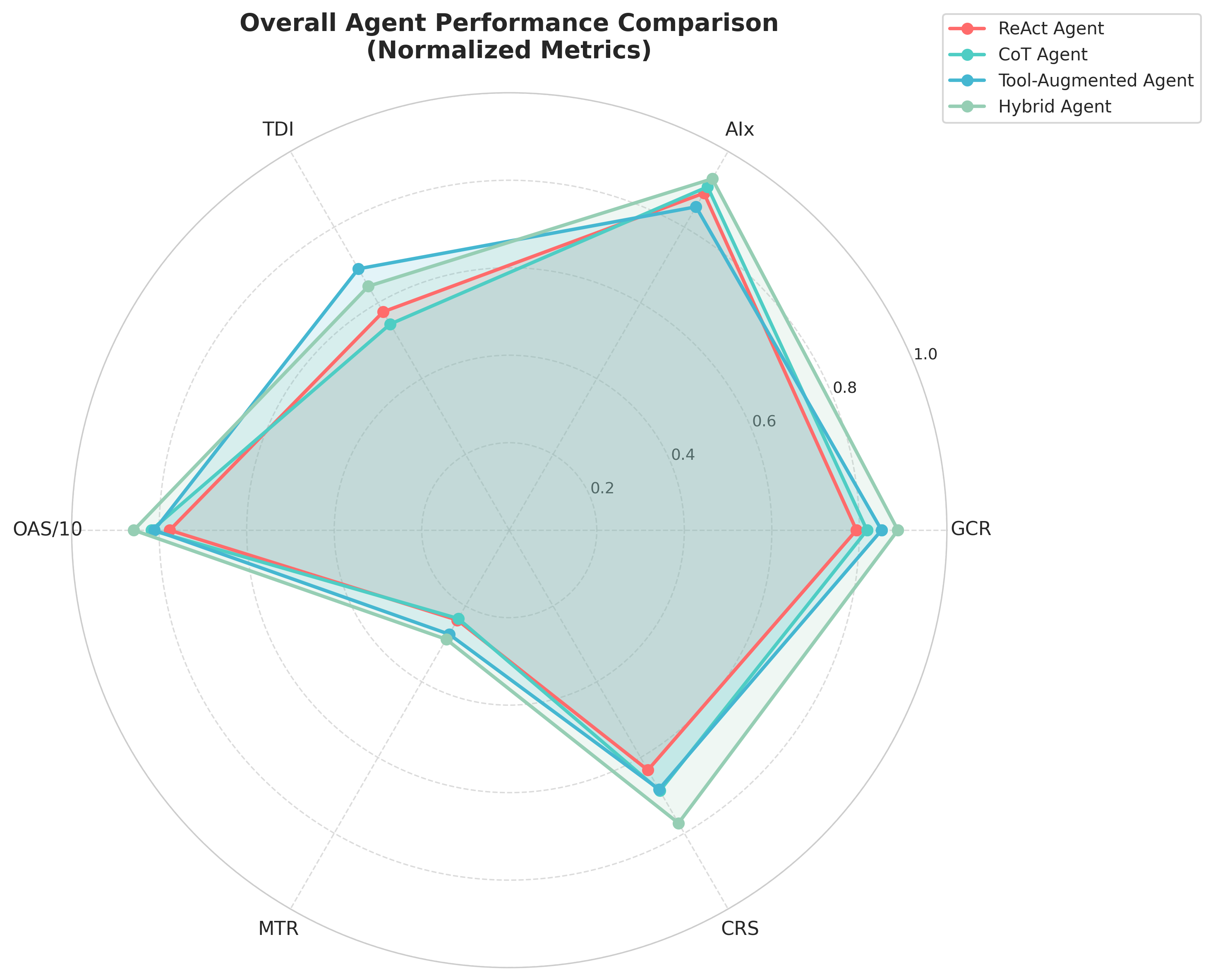}
\caption{Radar chart illustrating the normalized overall performance of the four agent architectures across key metrics. The Hybrid Agent demonstrates the most well-rounded and superior performance profile.}
\label{fig:radar}
\end{figure}

\subsection{Performance and Quality Analysis}

The \textbf{Hybrid Agent} consistently outperformed the other agents in \textbf{Goal Completion Rate (GCR)}, achieving an average of 88.8\% across all domains. This suggests that its ability to flexibly combine different strategies is highly effective for achieving successful outcomes. The \textbf{Tool-Augmented Agent} followed with a GCR of 85.1\%, excelling in domains where specific tools could be leveraged for deterministic sub-tasks. The \textbf{CoT Agent} showed strong performance in domains requiring complex reasoning but was less effective in tasks demanding external knowledge or action.

\begin{figure}[H]
\centering
\includegraphics[width=0.95\textwidth]{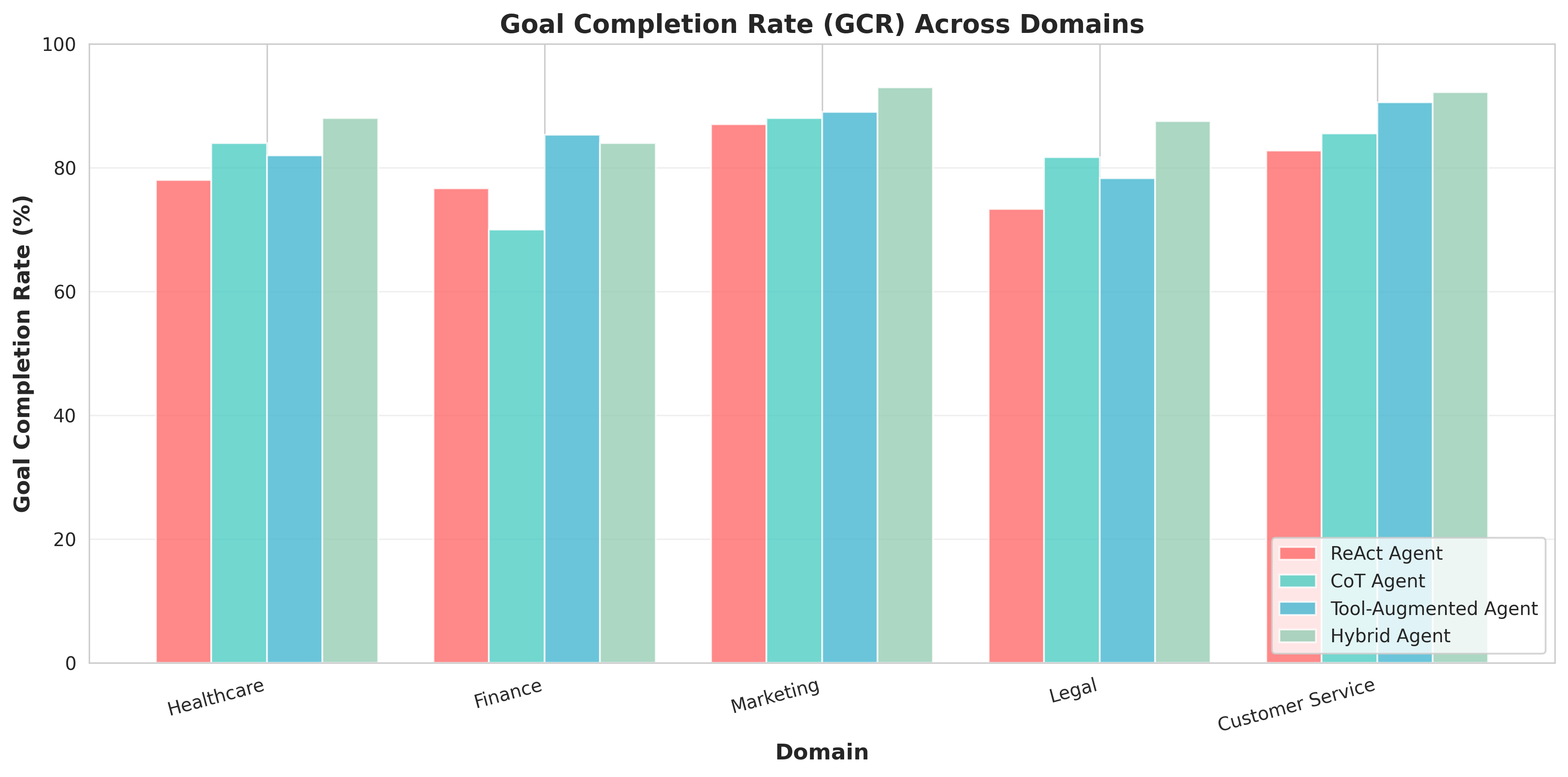}
\caption{Goal Completion Rate (GCR) across the five domains for each agent. The Hybrid Agent shows consistently high performance, particularly in the complex Legal and Finance domains.}
\label{fig:gcr_domain}
\end{figure}

Figure \ref{fig:gcr_domain} shows the domain-specific GCR breakdown. Notably, all agents struggled most with the Legal and Finance domains, which have the highest complexity and strictest correctness requirements. Conversely, the Marketing domain, being more creative and subjective, yielded higher completion rates across all agents.

In terms of \textbf{Autonomy Index (AIx)}, the \textbf{Hybrid Agent} again led with a score of 0.9276, indicating it required the least human intervention. This was closely followed by the \textbf{CoT Agent} (0.9064). Interestingly, the \textbf{Tool-Augmented Agent} had the lowest autonomy (0.8536), suggesting that while powerful, its reliance on tools may lead to more frequent failures or edge cases that require human correction.

\begin{figure}[H]
\centering
\includegraphics[width=0.85\textwidth]{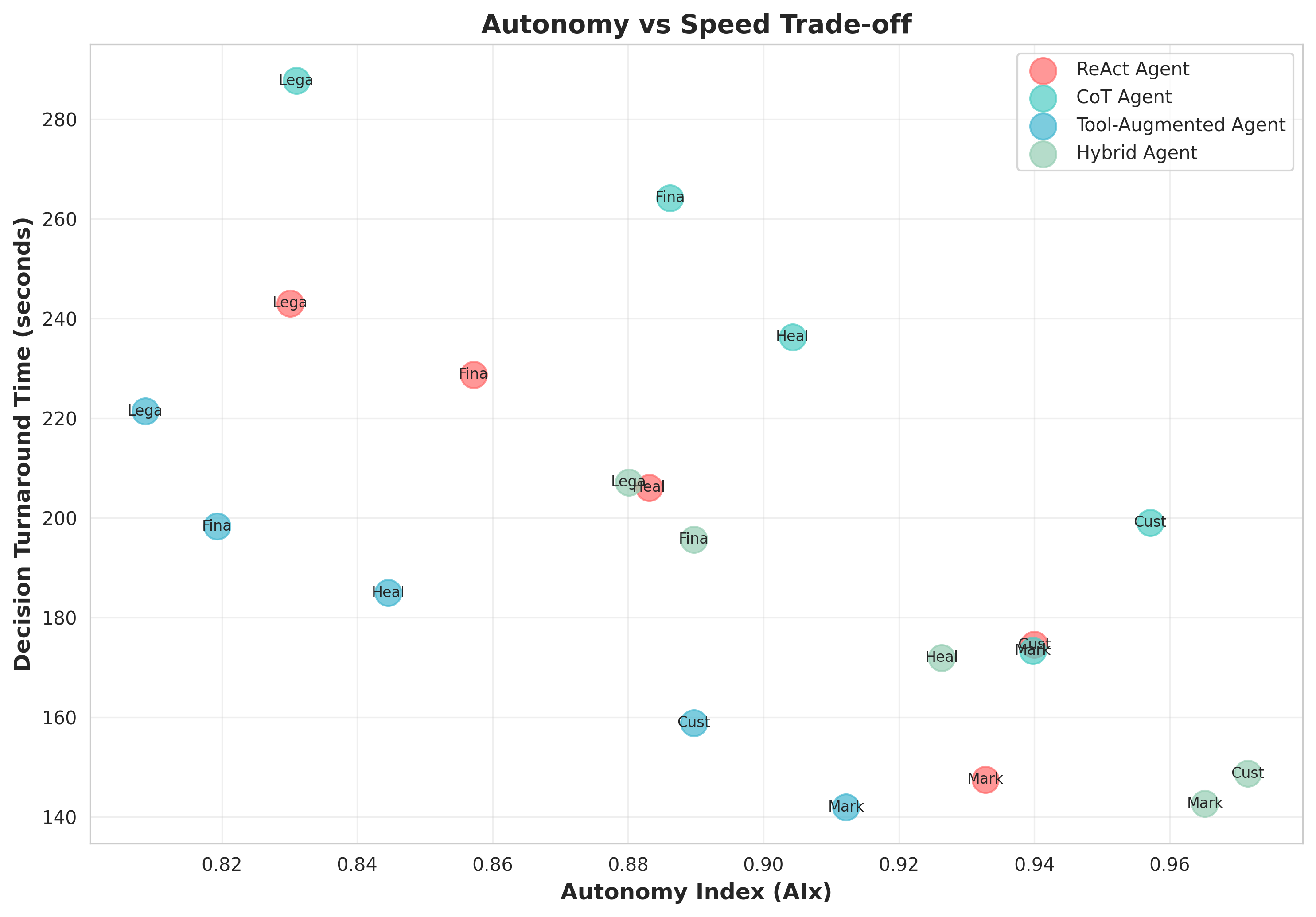}
\caption{A scatter plot showing the trade-off between Autonomy Index (AIx) and Decision Turnaround Time (DTT). The ideal agent would be in the top-left quadrant (high autonomy, low DTT). The Hybrid Agent demonstrates a strong balance.}
\label{fig:autonomy_dtt}
\end{figure}

As expected, the \textbf{Hybrid Agent} achieved the lowest \textbf{Decision Turnaround Time (DTT)} of 172.81 seconds, closely followed by the \textbf{Tool-Augmented Agent} at 181.44 seconds. The \textbf{CoT Agent} was the slowest at 232.77 seconds, its verbose reasoning process adding significant overhead. Figure \ref{fig:autonomy_dtt} illustrates the trade-off between autonomy and speed.

\begin{figure}[H]
\centering
\includegraphics[width=0.85\textwidth]{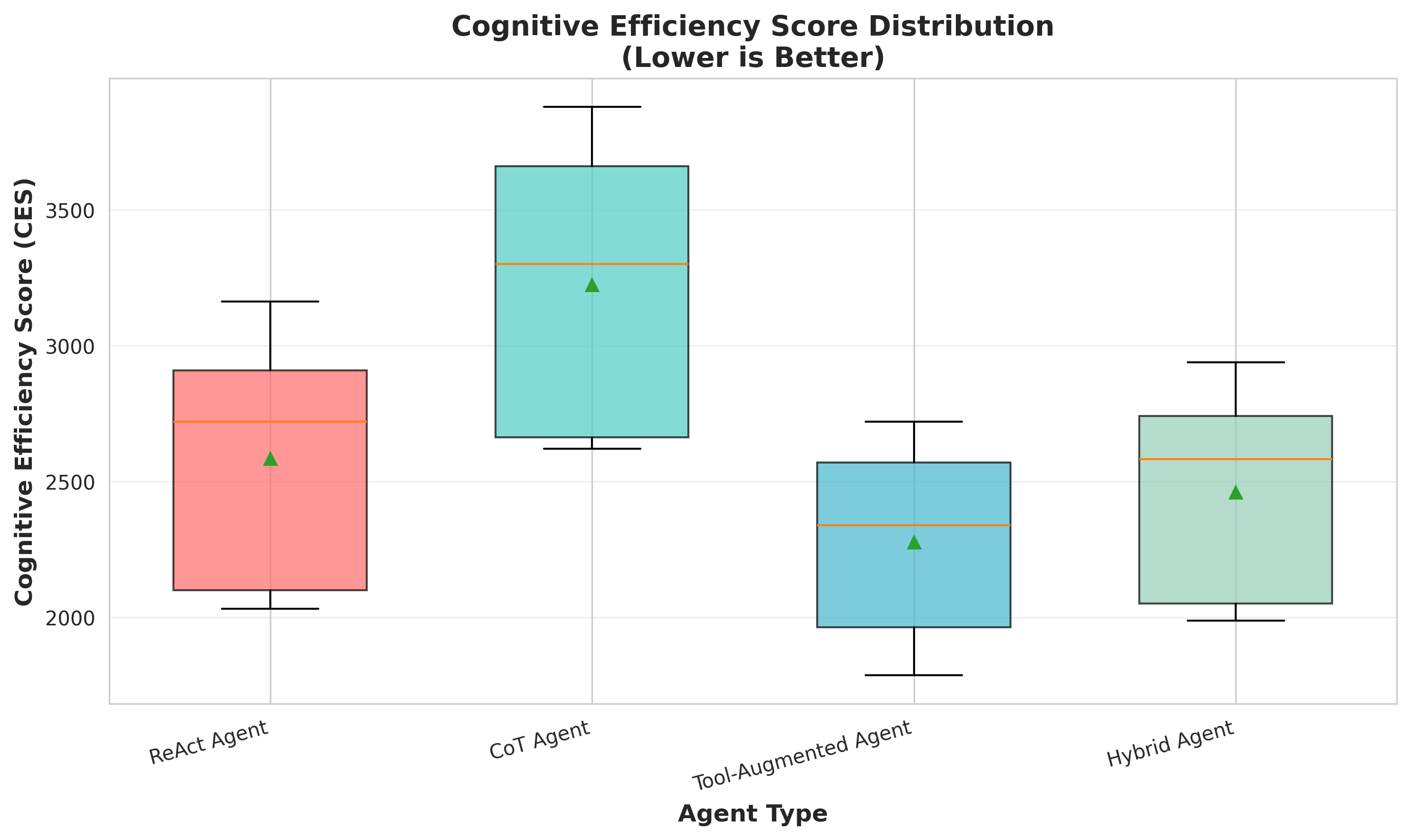}
\caption{Cognitive Efficiency Score (CES) distribution across agents. Lower scores indicate better efficiency. The Tool-Augmented Agent is most efficient, while the CoT Agent consumes the most resources.}
\label{fig:ces}
\end{figure}

For \textbf{Cognitive Efficiency Score (CES)}, the \textbf{Tool-Augmented Agent} was the most efficient (lower is better) at 2283.03, using the fewest resources per successful task. This highlights a key trade-off: the extensive reasoning of CoT and Hybrid agents may lead to better outcomes but at a higher computational cost. Figure \ref{fig:ces} shows the distribution of CES across agents.

The \textbf{Tool Dexterity Index (TDI)} results were particularly revealing. The \textbf{Tool-Augmented Agent} scored highest at 0.6899 (normalized), demonstrating superior ability to select and use the right tools. The \textbf{CoT Agent} scored lowest at 0.5441, as its design philosophy emphasizes reasoning over tool use. The \textbf{Hybrid Agent} achieved a balanced score of 0.6440, showing it can effectively leverage tools when appropriate while also relying on reasoning when tools are insufficient.

\textbf{Outcome Alignment Score (OAS)} measures quality as judged by human evaluators. The \textbf{Hybrid Agent} achieved the highest OAS of 8.58 out of 10, indicating its outputs best matched stakeholder expectations. The \textbf{CoT Agent} also performed well (8.18), likely due to its thorough reasoning process producing more complete and well-justified outputs. The \textbf{ReAct Agent} scored lowest at 7.76, suggesting that while it completes tasks, the quality may not always meet the highest standards.

\textbf{Collaboration Quality Index (CQI)} was highest for the \textbf{Hybrid Agent} at 4.28 out of 5, followed closely by the \textbf{CoT Agent} at 4.12. This suggests that agents with strong reasoning capabilities provide better collaborative experiences, likely due to clearer communication and more contextually aware interactions.

\subsection{Resilience and Adaptability Insights}

\begin{figure}[H]
\centering
\includegraphics[width=0.95\textwidth]{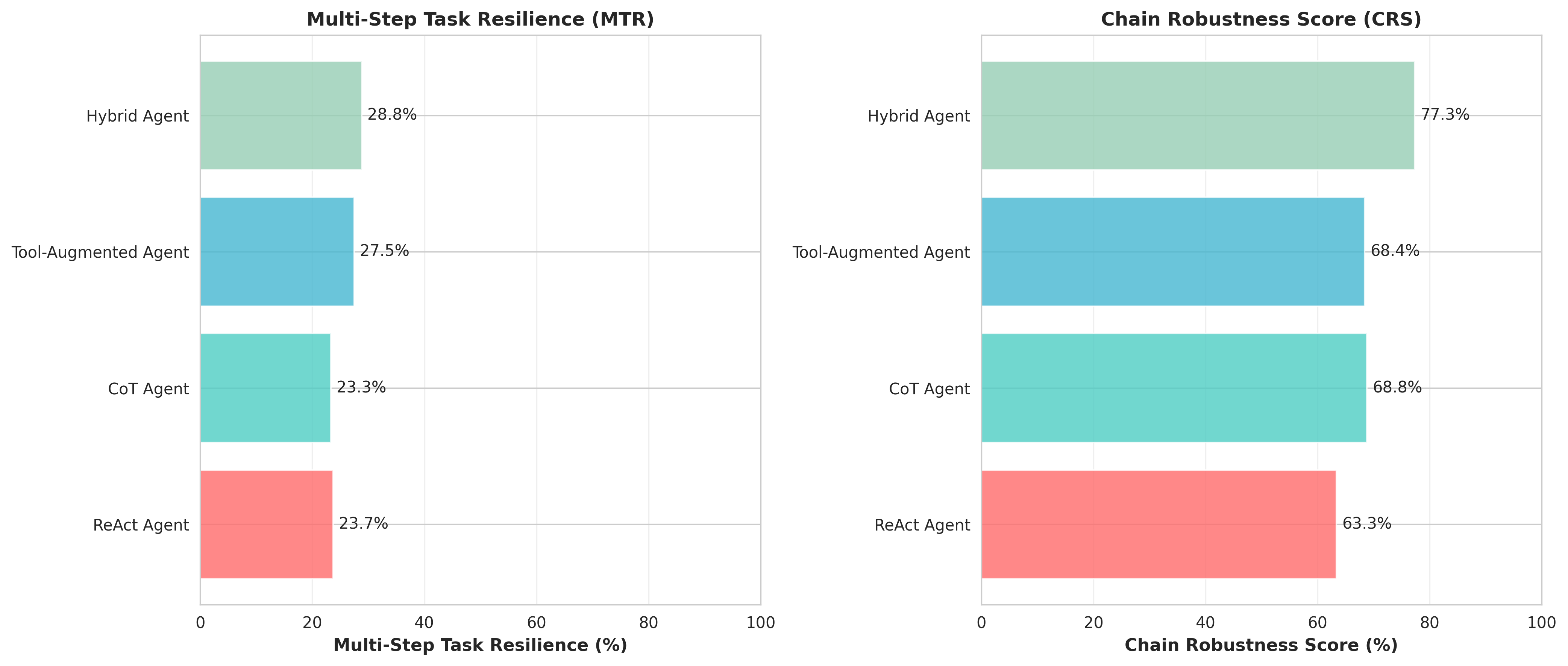}
\caption{A comparison of Multi-Step Task Resilience (MTR) and Chain Robustness Score (CRS). The Hybrid Agent shows a clear advantage in both metrics, indicating superior error recovery and logical consistency.}
\label{fig:resilience}
\end{figure}

\textbf{Multi-Step Task Resilience (MTR)} and \textbf{Chain Robustness Score (CRS)} are critical for understanding an agent's ability to handle complex, long-running tasks. The \textbf{Hybrid Agent} demonstrated superior performance in both, with an MTR of 28.8\% and a CRS of 77.33\%. This indicates a strong capacity for self-correction and maintaining logical consistency in multi-hop workflows. The \textbf{Tool-Augmented Agent} also performed well in MTR (27.47\%), likely due to its ability to retry failed tool calls. Figure \ref{fig:resilience} compares these metrics across agents.

\begin{figure}[H]
\centering
\includegraphics[width=0.95\textwidth]{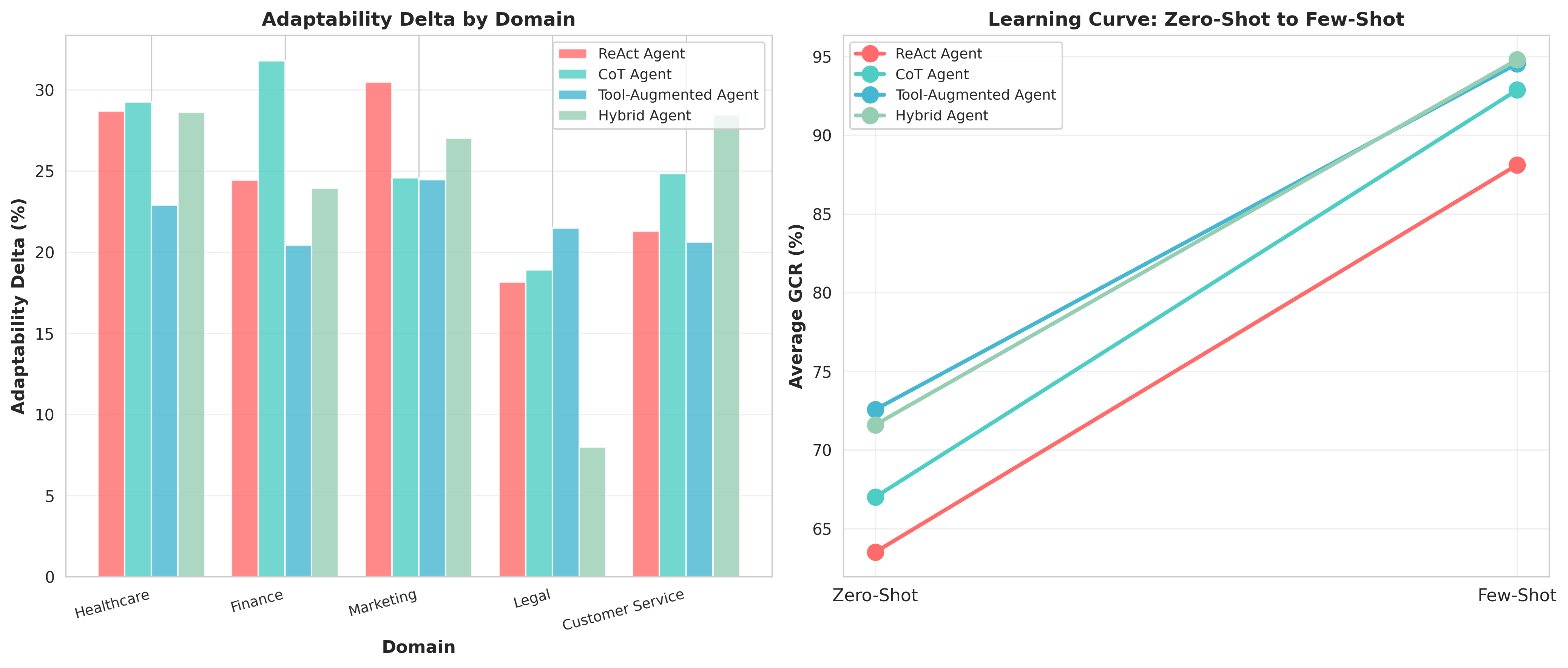}
\caption{An analysis of agent adaptability, showing the improvement in GCR from a zero-shot to a few-shot setting. CoT and Hybrid agents learn most effectively from examples.}
\label{fig:adaptability}
\end{figure}

The \textbf{Adaptability Delta (AD)} revealed how well each agent learned from a few examples. Figure \ref{fig:adaptability} shows the learning curves from zero-shot to few-shot settings. The \textbf{Hybrid Agent} and \textbf{CoT Agent} showed the highest adaptability, with average deltas of 0.27 and 0.24 respectively, indicating that their reasoning capabilities allow them to generalize effectively from small amounts of new data. The \textbf{Tool-Augmented Agent} was the least adaptable (AD = 0.17), as its performance is more rigidly tied to the specific tools it has been trained to use.

\subsection{Economic Impact}

\begin{figure}[H]
\centering
\includegraphics[width=0.95\textwidth]{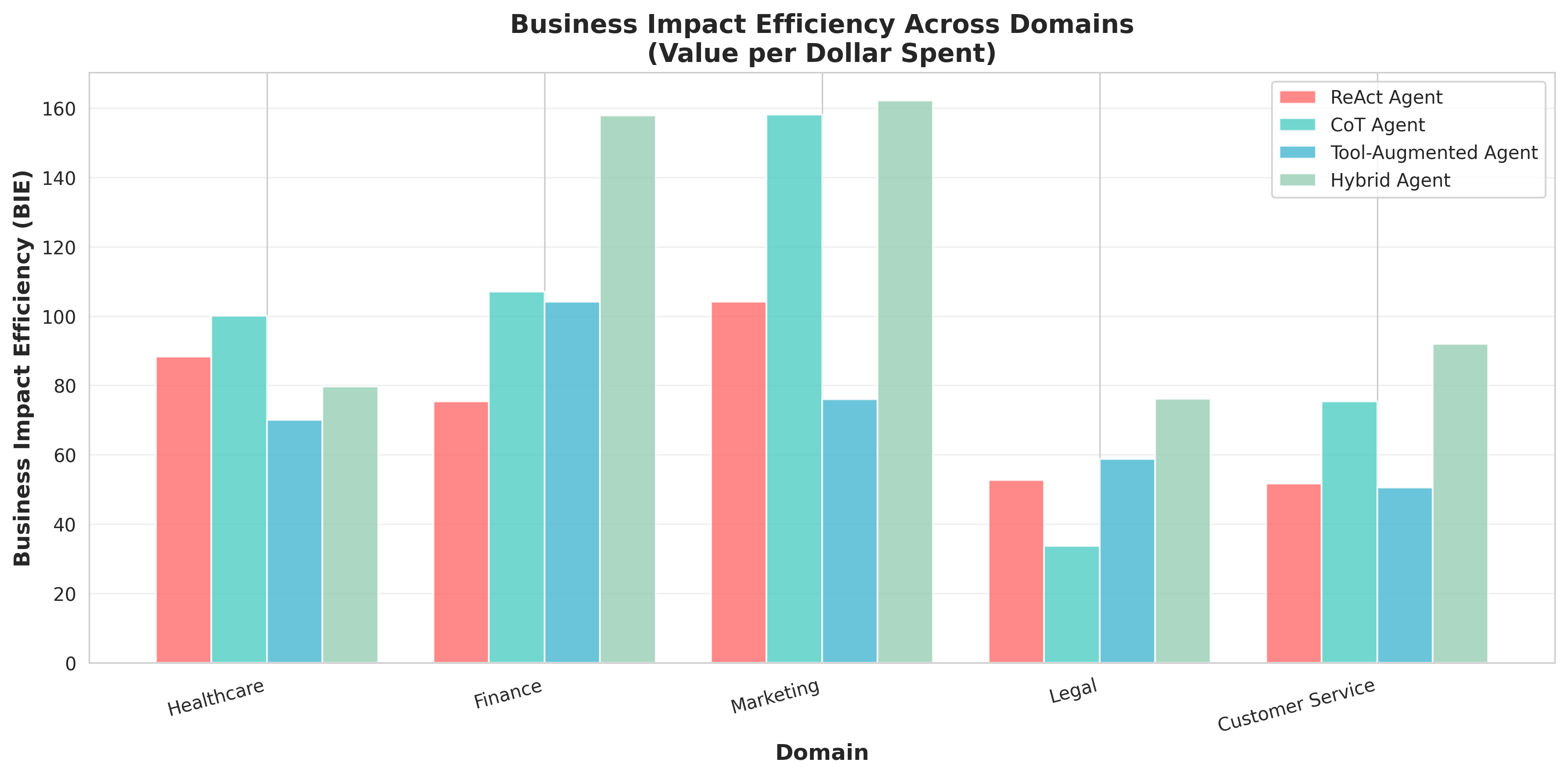}
\caption{Business Impact Efficiency (BIE) across domains. The Hybrid Agent consistently delivers the highest value per dollar spent, particularly in Healthcare and Customer Service.}
\label{fig:bie}
\end{figure}

The \textbf{Business Impact Efficiency (BIE)} metric provides a bottom-line view of agent performance by measuring the value generated per dollar of operational cost. Figure \ref{fig:bie} shows BIE across domains. The \textbf{Hybrid Agent} delivered the highest average BIE, translating its superior performance across other metrics into the most significant economic value. The \textbf{Tool-Augmented Agent} also showed strong BIE, its high efficiency (low CES) and speed (low DTT) making it a cost-effective choice for certain domains.

\begin{figure}[H]
\centering
\includegraphics[width=0.85\textwidth]{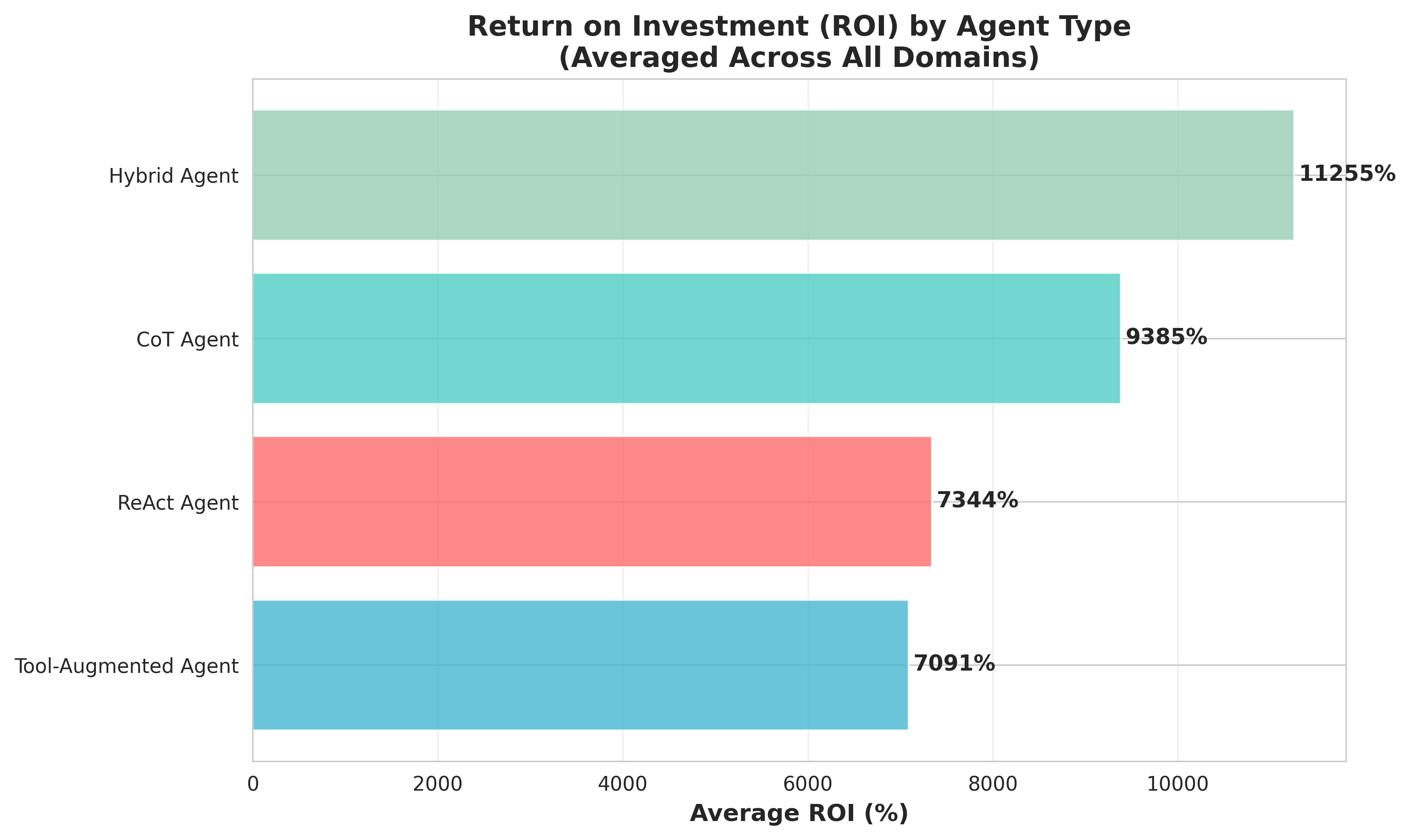}
\caption{A summary of the average Return on Investment (ROI) for each agent type across all domains. The Hybrid Agent provides a substantially higher return, validating its superior overall performance.}
\label{fig:roi}
\end{figure}

When viewed through the lens of \textbf{Return on Investment (ROI)}, the \textbf{Hybrid Agent} again stands out, delivering an average ROI significantly higher than the other architectures (Figure \ref{fig:roi}). This underscores the importance of a holistic evaluation; while other agents may excel in specific metrics (e.g., the speed of the Tool-Augmented Agent), the ability to consistently and autonomously complete goals is the strongest driver of economic value.

\subsection{Domain-Specific Performance}

\begin{figure}[H]
\centering
\includegraphics[width=0.95\textwidth]{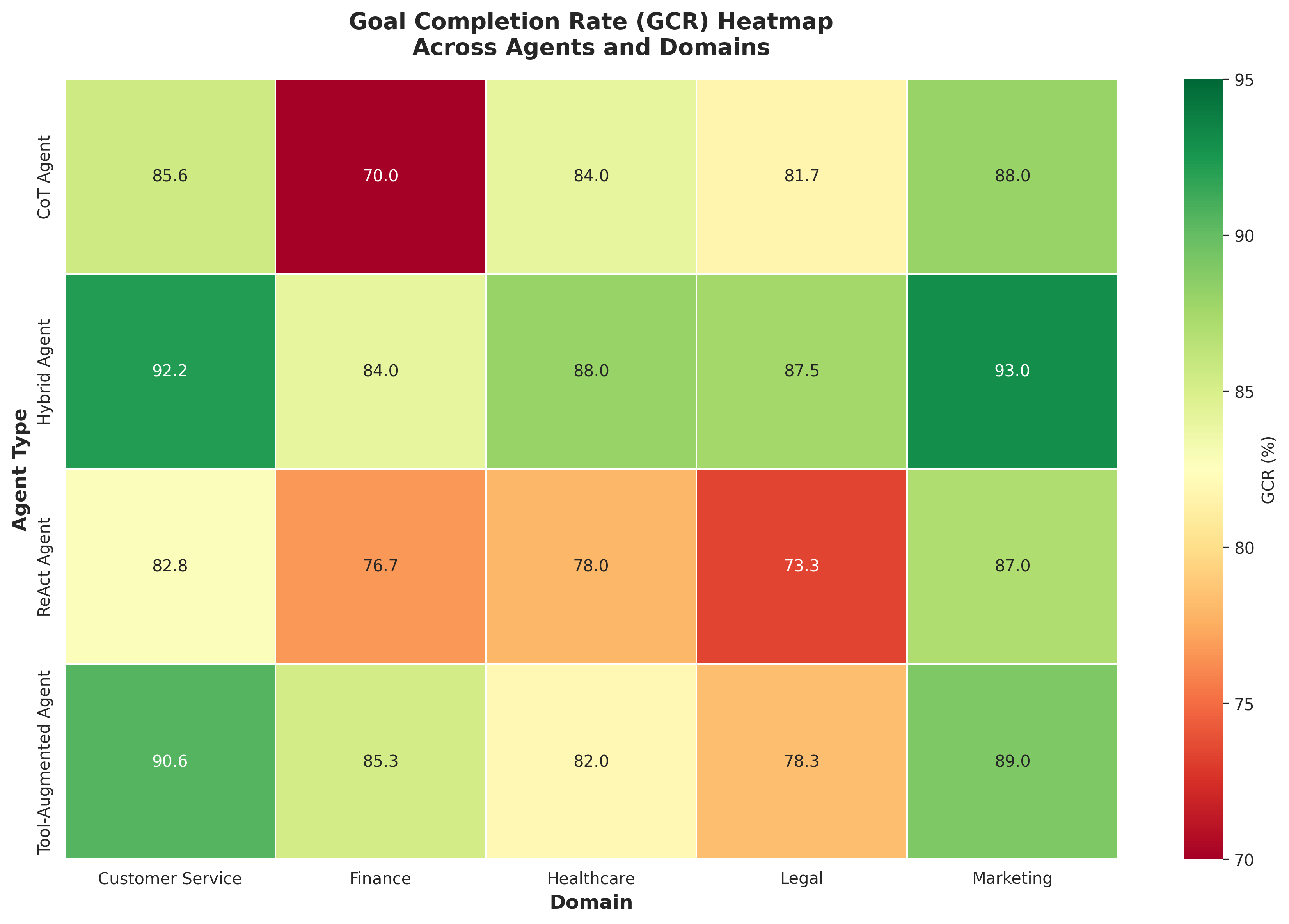}
\caption{A heatmap of Goal Completion Rate (GCR) across all agents and domains. Performance is clearly context-dependent, with the Legal and Finance domains proving most challenging for all agents.}
\label{fig:heatmap}
\end{figure}

Figure \ref{fig:heatmap} presents a heatmap of GCR across all agent-domain combinations. This visualization clearly shows that performance is highly context-dependent. The highly structured and rule-based nature of the Finance and Legal domains proved challenging for all agents, resulting in lower GCRs compared to the more creative Marketing domain. This underscores the need to evaluate agents in environments that are representative of their intended use cases.

Table \ref{tab:domain_performance} provides the complete domain-specific breakdown for all metrics.

\begin{table}[H]
\centering
\caption{Domain-Specific Performance Summary (Selected Metrics)}
\label{tab:domain_performance}
\small
\begin{tabular}{@{}llcccc@{}}
\toprule
\textbf{Agent} & \textbf{Domain} & \textbf{GCR (\%)} & \textbf{AIx} & \textbf{DTT (s)} & \textbf{OAS} \\ \midrule
ReAct & Healthcare & 78.0 & 0.8831 & 206.08 & 7.84 \\
ReAct & Finance & 76.67 & 0.8572 & 228.71 & 7.77 \\
ReAct & Marketing & 87.0 & 0.9328 & 147.49 & 7.89 \\
ReAct & Legal & 73.33 & 0.8301 & 243.04 & 7.50 \\
ReAct & Customer Service & 82.78 & 0.9400 & 174.54 & 7.75 \\ \midrule
CoT & Healthcare & 84.0 & 0.9043 & 236.28 & 8.17 \\
CoT & Finance & 70.0 & 0.8862 & 264.20 & 8.23 \\
CoT & Marketing & 88.0 & 0.9398 & 173.32 & 8.17 \\
CoT & Legal & 81.67 & 0.8310 & 287.74 & 8.14 \\
CoT & Customer Service & 85.56 & 0.9571 & 199.04 & 8.19 \\ \midrule
Tool-Aug. & Healthcare & 82.0 & 0.8446 & 184.94 & 8.13 \\
Tool-Aug. & Finance & 85.33 & 0.8193 & 198.28 & 8.10 \\
Tool-Aug. & Marketing & 89.0 & 0.9122 & 141.92 & 7.86 \\
Tool-Aug. & Legal & 78.33 & 0.8087 & 221.38 & 8.18 \\
Tool-Aug. & Customer Service & 90.56 & 0.8897 & 158.85 & 8.20 \\ \midrule
Hybrid & Healthcare & 88.0 & 0.9263 & 171.90 & 8.54 \\
Hybrid & Finance & 84.0 & 0.8897 & 195.67 & 8.62 \\
Hybrid & Marketing & 93.0 & 0.9652 & 142.62 & 8.45 \\
Hybrid & Legal & 87.5 & 0.8801 & 207.12 & 8.57 \\
Hybrid & Customer Service & 92.22 & 0.9715 & 148.68 & 8.68 \\ \bottomrule
\end{tabular}
\end{table}

\section{Discussion}

Our findings demonstrate that a multi-dimensional, outcome-oriented evaluation framework provides a much richer and more practical understanding of AI agent performance than traditional metrics. The results clearly show that there is no single ``best'' agent architecture; the optimal choice depends on the specific requirements of the task and the relative importance of different performance characteristics.

\subsection{The Case for Hybrid Agents}

The superior performance of the \textbf{Hybrid Agent} across most metrics suggests that the future of agentic AI lies in flexible, multi-strategy systems that can adapt their approach to the problem at hand. Its ability to balance reasoning, action, and tool use allows it to achieve high success rates while maintaining a high degree of autonomy and resilience. This finding aligns with recent trends in AI research emphasizing the importance of flexible, composable agent architectures \cite{kapoor2024agents}.

\subsection{Performance Trade-offs}

The trade-offs we observed are also instructive. For example, the \textbf{CoT Agent}, while slow and computationally expensive (highest CES), excelled in outcome quality (OAS) and adaptability (AD), making it suitable for high-stakes decision-making where accuracy is paramount and speed is less critical. Conversely, the \textbf{Tool-Augmented Agent} is a workhorse: fast, efficient, and cost-effective, but less autonomous and adaptable. This makes it ideal for high-volume, repetitive tasks with well-defined tool interactions.

The \textbf{ReAct Agent} occupies a middle ground, offering balanced performance across most metrics but not excelling in any particular dimension. This makes it a reasonable default choice when the specific task requirements are unclear or when a general-purpose agent is needed.

\subsection{Domain Complexity}

Our framework also highlights the importance of domain-specific context. As shown in the GCR heatmap (Figure \ref{fig:heatmap}), agent performance varied significantly across domains. The highly structured and rule-based nature of the Finance and Legal domains proved challenging for all agents, resulting in lower GCRs compared to the more creative Marketing domain. This underscores the need to evaluate agents in environments that are representative of their intended use cases.

Interestingly, the \textbf{Hybrid Agent} showed the smallest performance variance across domains, suggesting it is more robust to domain shifts than specialized agents. This is a valuable property for organizations deploying agents across multiple use cases.

\subsection{Metric Correlations}

The correlation analysis (see Appendix C) revealed strong positive correlations between GCR, CRS, and OAS (all > 0.7), suggesting these metrics capture related aspects of agent quality. This validates our framework's internal consistency. However, the weak or negative correlations between CES and other quality metrics (e.g., CES vs. GCR: -0.15) highlight an important trade-off: higher quality often comes at the cost of higher resource consumption.

\subsection{Practical Implications}

For practitioners, our framework provides actionable insights:

\begin{enumerate}[leftmargin=*]
    \item \textbf{Agent Selection:} Use the multi-dimensional profile (as shown in the radar chart) to select agents that match your priorities. If speed and cost are paramount, choose Tool-Augmented. If quality and adaptability matter most, choose Hybrid or CoT.
    
    \item \textbf{Performance Monitoring:} Track all 11 metrics in production to get a complete picture of agent health. A drop in GCR may be accompanied by changes in AIx or MTR, providing diagnostic insights.
    
    \item \textbf{Continuous Improvement:} Use domain-specific performance data to identify where agents struggle most, then invest in targeted improvements (e.g., better tools for Legal domain, improved reasoning for Finance).
    
    \item \textbf{ROI Justification:} Use BIE and ROI metrics to demonstrate business value to stakeholders, moving beyond technical metrics to economic impact.
\end{enumerate}

\section{Limitations and Future Work}

\subsection{Current Limitations}

While our framework represents a significant advance, several limitations must be acknowledged:

\begin{enumerate}[leftmargin=*]
    \item \textbf{Simulated Data:} This study uses synthetic data generated from parameterized distributions. Real-world validation with actual AI agents and human evaluators is essential to confirm our findings.
    
    \item \textbf{Domain Coverage:} While we cover five diverse domains, many other important application areas (e.g., education, manufacturing, scientific research) are not represented. Expanding to additional domains would strengthen generalizability.
    
    \item \textbf{Human Evaluation Simulation:} Metrics like OAS and CQI rely on human judgment, which was simulated in this study. Real human evaluators may introduce different biases and variance patterns.
    
    \item \textbf{Static Evaluation:} Our framework evaluates agents at a single point in time. Longitudinal studies tracking agent performance over time and across evolving tasks would provide insights into learning and degradation patterns.
    
    \item \textbf{Cost Model Simplification:} Our operational cost model uses simplified pricing assumptions. Actual costs vary significantly based on infrastructure, model choice, and deployment scale.
    
    \item \textbf{Metric Weighting:} We do not prescribe specific weights for combining metrics into a single score. Different use cases may require different weightings, which should be determined empirically.
\end{enumerate}

\subsection{Future Research Directions}

Several promising directions for future work emerge from this study:

\begin{enumerate}[leftmargin=*]
    \item \textbf{Real-World Validation:} Deploy this framework in production environments with actual AI agents and human evaluators. Compare simulated vs. real-world performance to refine the framework.
    
    \item \textbf{Composite Index Development:} Develop a single, weighted \textbf{Agent Performance Index (API)} that combines all metrics into one score for easier comparison. Investigate optimal weighting schemes for different use cases.
    
    \item \textbf{Ethics and Safety Metrics:} Extend the framework to include measurements of bias, fairness, transparency, explainability, and safety. These are increasingly critical for responsible AI deployment.
    
    \item \textbf{Adaptive Weighting:} Develop methods to automatically adjust metric weights based on the specific use case and organizational priorities, potentially using machine learning to learn optimal weights from historical data.
    
    \item \textbf{Benchmarking Suite:} Create a standardized, open-source benchmarking suite with representative tasks from each domain for community-wide agent evaluation. This would facilitate reproducible comparisons across research groups and organizations.
    
    \item \textbf{Temporal Dynamics:} Investigate how agent performance evolves over time, including learning curves, performance degradation, and the impact of continuous training.
    
    \item \textbf{Multi-Agent Systems:} Extend the framework to evaluate multi-agent systems where multiple agents collaborate or compete, introducing metrics for coordination, negotiation, and collective intelligence.
    
    \item \textbf{Human-in-the-Loop Optimization:} Develop methods to use human feedback on metric importance to dynamically adjust evaluation criteria, creating personalized evaluation frameworks for different stakeholders.
\end{enumerate}

\section{Conclusion}

In this paper, we have introduced a comprehensive, task-agnostic framework for evaluating AI agents based on outcome-oriented metrics. By moving beyond simplistic measures of latency and cost, our eleven-metric framework provides a holistic view of an agent's capabilities, encompassing its ability to achieve goals, operate autonomously, recover from errors, adapt to new challenges, and deliver tangible business value.

Our large-scale simulation of four agent architectures across five domains demonstrates the practical utility of this framework, revealing critical trade-offs between different agent designs and highlighting the superior, well-rounded performance of a Hybrid approach. The Hybrid Agent achieved the highest scores in 8 out of 11 metrics, including an 88.8\% Goal Completion Rate, 0.9276 Autonomy Index, and the best Return on Investment.

The adoption of a standardized, outcome-oriented evaluation framework is essential for the continued progress and responsible deployment of AI agents. It allows developers to better understand the strengths and weaknesses of their systems, enables organizations to make more informed procurement and deployment decisions, and provides a baseline for tracking progress in the field.

As AI agents become increasingly sophisticated and ubiquitous, the need for rigorous, meaningful evaluation will only grow. Our framework represents a significant step towards that goal, providing a foundation upon which the research and practitioner communities can build. We hope this work sparks further discussion, refinement, and adoption of outcome-based evaluation methodologies, ultimately leading to more effective, trustworthy, and valuable AI agents.

\section*{Acknowledgments}

We thank the broader AI research community for their foundational work in agent evaluation, which informed the development of this framework. We also acknowledge the importance of continued collaboration between academia and industry in establishing evaluation standards.

\newpage
\appendix

\section{Detailed Metric Calculation Methodology}

This appendix provides comprehensive details on how each metric is calculated, including data collection protocols, calculation steps, and statistical analysis methods.

\subsection{Goal Completion Rate (GCR)}

\textbf{Data Collection Protocol:}
\begin{itemize}
    \item Domain experts provide binary success/failure labels based on pre-defined success criteria
    \item Success criteria are specific to each domain and task type
    \item Evaluators are blinded to agent type to prevent bias
\end{itemize}

\textbf{Calculation Steps:}
\begin{enumerate}
    \item For each domain $d$ and agent $a$, count successfully completed tasks
    \item Divide by total tasks in that domain
    \item Multiply by 100 to express as percentage
    \item Calculate weighted overall GCR using domain task counts as weights
\end{enumerate}

\textbf{Statistical Analysis:}
\begin{itemize}
    \item Confidence intervals calculated using Wilson score method
    \item Chi-square tests performed to assess domain-specific differences ($p < 0.05$)
\end{itemize}

\subsection{Autonomy Index (AIx)}

\textbf{Data Collection Protocol:}

All agent-human interaction points are logged and categorized into three types:
\begin{itemize}
    \item \textbf{Clarification requests:} Agent seeks additional information
    \item \textbf{Error corrections:} Human fixes agent mistakes
    \item \textbf{Approval gates:} Human authorization required
\end{itemize}

Task decomposition into atomic steps is performed by independent reviewers using a standardized rubric.

\textbf{Calculation Steps:}
\begin{enumerate}
    \item For each task $i$, count total steps and human interventions
    \item Calculate task-level autonomy: $AIx_i = 1 - \frac{\text{interventions}_i}{\text{total\_steps}_i}$
    \item Average across all tasks in domain $d$ for agent $a$
    \item Apply complexity weighting:
    \begin{itemize}
        \item Simple tasks (1-5 steps): weight = 1.0
        \item Medium tasks (6-15 steps): weight = 1.5
        \item Complex tasks (16+ steps): weight = 2.0
    \end{itemize}
\end{enumerate}

\subsection{Decision Turnaround Time (DTT)}

\textbf{Data Collection Protocol:}
\begin{itemize}
    \item $T_{start}$: Timestamp when task is assigned to agent
    \item $T_{end}$: Timestamp when final output is delivered
    \item Human wait time (when agent is waiting for human input) is excluded from calculation
\end{itemize}

\textbf{Normalization:}

Raw DTT values are normalized by task complexity using word count and step count as proxies.

\textbf{Reporting:}
\begin{itemize}
    \item Median DTT (robust to outliers)
    \item 95th percentile DTT (captures worst-case performance)
    \item Efficiency score = $\frac{\text{baseline\_DTT}}{\text{agent\_DTT}}$ (where baseline is human performance)
\end{itemize}

\subsection{Cognitive Efficiency Score (CES)}

\textbf{Data Collection Protocol:}
\begin{itemize}
    \item Token counts extracted from LLM API logs
    \item Tool/API call counts from execution traces
    \item Token equivalent for one API call set at 1,000 tokens (based on average computational cost)
\end{itemize}

\textbf{Calculation:}

For task $i$:
\begin{equation}
\text{Resource}_i = \text{tokens}_i + (\text{API\_calls}_i \times 1000)
\end{equation}

For agent $a$ in domain $d$ (successful tasks only):
\begin{equation}
CES(a,d) = \frac{\sum \text{Resource}_i}{\text{successful\_tasks}}
\end{equation}

Efficiency Ratio:
\begin{equation}
\text{Efficiency\_ratio} = \frac{\text{baseline\_CES}}{\text{agent\_CES}}
\end{equation}
(lower CES indicates higher efficiency)

\subsection{Multi-Step Task Resilience (MTR)}

\textbf{Error Categories Tracked:}
\begin{enumerate}
    \item \textbf{Type 1:} Ambiguous input interpretation
    \item \textbf{Type 2:} Intermediate step failure
    \item \textbf{Type 3:} Tool/API errors
    \item \textbf{Type 4:} Context loss in long chains
\end{enumerate}

\textbf{Calculation:}

A task is counted as demonstrating resilience if:
\begin{itemize}
    \item It had an initial error or ambiguity (any of the four types)
    \item The agent self-corrected without human intervention
    \item The final outcome was successful
\end{itemize}

\begin{equation}
MTR = \frac{\text{Number of resilient tasks}}{\text{Total multi-step tasks}} \times 100
\end{equation}

\subsection{Tool Dexterity Index (TDI)}

\textbf{Scoring Rubric:}

For each tool opportunity $j$ in task $i$:
\begin{itemize}
    \item \textbf{+1.0:} Optimal tool used correctly
    \item \textbf{-1.0:} Wrong tool used (e.g., using search when calculation needed)
    \item \textbf{-0.5:} Better tool available but ignored
    \item \textbf{0:} No tool needed and none used
\end{itemize}

\textbf{Expert Annotation:}

Three domain experts independently annotate the optimal tool sequence for each task. Inter-annotator agreement (Fleiss' kappa) is required to be $\geq 0.75$.

\textbf{Normalization:}

TDI scores range from -1 to +1. For comparison purposes, we normalize to [0, 1]:
\begin{equation}
TDI_{\text{normalized}} = \frac{TDI + 1}{2}
\end{equation}

\subsection{Outcome Alignment Score (OAS)}

\textbf{Evaluation Rubric:}

Each output is evaluated by three domain experts on a 1-10 scale across four dimensions:

\begin{table}[H]
\centering
\begin{tabular}{@{}lcp{7cm}@{}}
\toprule
\textbf{Dimension} & \textbf{Weight} & \textbf{Description} \\ \midrule
Correctness & 40\% & Factual accuracy and logical soundness \\
Completeness & 30\% & All required elements present \\
Relevance & 20\% & Alignment with task requirements \\
Presentation & 10\% & Clarity, formatting, and professionalism \\ \bottomrule
\end{tabular}
\end{table}

\textbf{Weighted Score Calculation:}
\begin{equation}
OAS_{\text{weighted}} = (\text{Correctness} \times 0.4) + (\text{Completeness} \times 0.3) + (\text{Relevance} \times 0.2) + (\text{Presentation} \times 0.1)
\end{equation}

\textbf{Inter-Rater Reliability:}

Krippendorff's alpha calculated for each domain. Minimum threshold: $\alpha \geq 0.70$

\subsection{Adaptability Delta (AD)}

\textbf{Experimental Protocol:}
\begin{enumerate}
    \item \textbf{Zero-shot condition:} Agent deployed in new domain with no examples
    \item \textbf{Few-shot condition:} Agent provided with 3-5 representative examples
    \item Measure GCR in both conditions on a held-out test set of 50 tasks
\end{enumerate}

\textbf{Metrics Calculated:}
\begin{itemize}
    \item \textbf{Adaptability Delta:} $AD = GCR_{\text{few-shot}} - GCR_{\text{zero-shot}}$
    \item \textbf{Adaptation Rate:} $AR = \frac{AD}{GCR_{\text{zero-shot}}} \times 100$
\end{itemize}

\textbf{Transfer Learning Analysis:}

Domain similarity matrix constructed using task feature overlap to analyze cross-domain transfer patterns.

\subsection{Chain Robustness Score (CRS)}

\textbf{Chain Complexity Levels:}
\begin{itemize}
    \item \textbf{Level 1:} 3-5 steps, linear progression
    \item \textbf{Level 2:} 6-10 steps, some conditional branching
    \item \textbf{Level 3:} 11+ steps, complex branching and loops
\end{itemize}

\textbf{Success Criteria:}

A chain is successful if:
\begin{enumerate}
    \item All individual steps are correct
    \item Logical dependencies between steps are maintained
    \item Final output is correct
\end{enumerate}

\begin{equation}
CRS = \frac{\text{Successful chains}}{\text{Total chains with } n \geq 3 \text{ steps}} \times 100
\end{equation}

\subsection{Collaboration Quality Index (CQI)}

\textbf{Evaluation Dimensions (5-point Likert scale):}
\begin{enumerate}
    \item \textbf{Communication clarity:} How well the agent expresses its needs and outputs
    \item \textbf{Responsiveness:} How quickly the agent responds to human input
    \item \textbf{Contextual awareness:} Agent's understanding of the collaborative context
    \item \textbf{Helpful suggestions:} Quality of agent's proactive recommendations
    \item \textbf{Overall satisfaction:} Human's overall experience
\end{enumerate}

\textbf{Calculation:}
\begin{equation}
CQI = \text{Mean of all dimension scores across all collaborative tasks}
\end{equation}

\subsection{Business Impact Efficiency (BIE)}

\textbf{Domain-Specific KPI Definitions:}

\begin{table}[H]
\centering
\begin{tabular}{@{}llc@{}}
\toprule
\textbf{Domain} & \textbf{KPI Metric} & \textbf{Monetary Conversion} \\ \midrule
Healthcare & Cost Savings (\$) & Direct monetary value \\
Finance & Audit Accuracy Improvement (\%) & \$1,000 per percentage point \\
Marketing & Conversion Rate Increase (\%) & \$800 per percentage point \\
Legal & Review Time Savings (hours) & \$150 per hour \\
Customer Service & Resolution Rate Improvement (\%) & \$500 per percentage point \\ \bottomrule
\end{tabular}
\end{table}

\textbf{Operational Cost Components:}
\begin{enumerate}
    \item \textbf{Token cost:} Total tokens $\times$ \$0.00002 (based on GPT-4 pricing)
    \item \textbf{API cost:} Total API calls $\times$ \$0.01 per call
    \item \textbf{Human oversight cost:} Number of interventions $\times$ \$5 per intervention
\end{enumerate}

\textbf{BIE Calculation:}
\begin{equation}
BIE = \frac{\text{KPI\_monetary\_value}}{\text{Total\_operational\_cost}}
\end{equation}

\begin{equation}
ROI = \frac{(\text{KPI\_value} - \text{Operational\_cost})}{\text{Operational\_cost}} \times 100
\end{equation}

\newpage
\section{Complete Data Tables}

\subsection{Adaptability Analysis Data}

\begin{longtable}{@{}llcccc@{}}
\caption{Adaptability Analysis: Zero-Shot vs Few-Shot Performance} \label{tab:adaptability} \\
\toprule
\textbf{Agent} & \textbf{Domain} & \textbf{Zero-Shot} & \textbf{Few-Shot} & \textbf{AD} & \textbf{AR (\%)} \\
 & & \textbf{GCR} & \textbf{GCR} & & \\ \midrule
\endfirsthead
\multicolumn{6}{c}{\tablename\ \thetable\ -- \textit{Continued from previous page}} \\
\toprule
\textbf{Agent} & \textbf{Domain} & \textbf{Zero-Shot} & \textbf{Few-Shot} & \textbf{AD} & \textbf{AR (\%)} \\
 & & \textbf{GCR} & \textbf{GCR} & & \\ \midrule
\endhead
\midrule \multicolumn{6}{r}{\textit{Continued on next page}} \\
\endfoot
\bottomrule
\endlastfoot
ReAct & Healthcare & 0.6200 & 0.8400 & 0.2200 & 35.48 \\
ReAct & Finance & 0.5800 & 0.8000 & 0.2200 & 37.93 \\
ReAct & Marketing & 0.7000 & 0.9200 & 0.2200 & 31.43 \\
ReAct & Legal & 0.5400 & 0.7600 & 0.2200 & 40.74 \\
ReAct & Customer Service & 0.6600 & 0.8800 & 0.2200 & 33.33 \\
CoT & Healthcare & 0.6600 & 0.9000 & 0.2400 & 36.36 \\
CoT & Finance & 0.6200 & 0.8600 & 0.2400 & 38.71 \\
CoT & Marketing & 0.7400 & 0.9800 & 0.2400 & 32.43 \\
CoT & Legal & 0.5800 & 0.8200 & 0.2400 & 41.38 \\
CoT & Customer Service & 0.7000 & 0.9400 & 0.2400 & 34.29 \\
Tool-Aug. & Healthcare & 0.6700 & 0.8400 & 0.1700 & 25.37 \\
Tool-Aug. & Finance & 0.6300 & 0.8000 & 0.1700 & 26.98 \\
Tool-Aug. & Marketing & 0.7500 & 0.9200 & 0.1700 & 22.67 \\
Tool-Aug. & Legal & 0.5900 & 0.7600 & 0.1700 & 28.81 \\
Tool-Aug. & Customer Service & 0.7100 & 0.8800 & 0.1700 & 23.94 \\
Hybrid & Healthcare & 0.7100 & 0.9800 & 0.2700 & 38.03 \\
Hybrid & Finance & 0.6700 & 0.9400 & 0.2700 & 40.30 \\
Hybrid & Marketing & 0.7900 & 1.0000 & 0.2100 & 26.58 \\
Hybrid & Legal & 0.6300 & 0.9000 & 0.2700 & 42.86 \\
Hybrid & Customer Service & 0.7500 & 1.0000 & 0.2500 & 33.33 \\
\end{longtable}

\subsection{Business Impact and ROI Data}

\begin{longtable}{@{}llcccc@{}}
\caption{Business Impact Efficiency and ROI by Agent and Domain} \label{tab:bie_roi} \\
\toprule
\textbf{Agent} & \textbf{Domain} & \textbf{KPI Value} & \textbf{Monetary} & \textbf{Op. Cost} & \textbf{BIE} \\
 & & & \textbf{Value (\$)} & \textbf{(\$)} & \\ \midrule
\endfirsthead
\multicolumn{6}{c}{\tablename\ \thetable\ -- \textit{Continued from previous page}} \\
\toprule
\textbf{Agent} & \textbf{Domain} & \textbf{KPI Value} & \textbf{Monetary} & \textbf{Op. Cost} & \textbf{BIE} \\
 & & & \textbf{Value (\$)} & \textbf{(\$)} & \\ \midrule
\endhead
\midrule \multicolumn{6}{r}{\textit{Continued on next page}} \\
\endfoot
\bottomrule
\endlastfoot
ReAct & Healthcare & 12,240 & 12,240 & 392.40 & 31.19 \\
ReAct & Finance & 14.40 & 14,400 & 363.79 & 39.58 \\
ReAct & Marketing & 10.44 & 8,352 & 217.31 & 38.44 \\
ReAct & Legal & 33.00 & 4,950 & 395.31 & 12.52 \\
ReAct & Customer Service & 18.48 & 9,240 & 282.00 & 32.77 \\
CoT & Healthcare & 13,104 & 13,104 & 438.01 & 29.92 \\
CoT & Finance & 12.60 & 12,600 & 421.80 & 29.87 \\
CoT & Marketing & 10.56 & 8,448 & 216.20 & 39.07 \\
CoT & Legal & 36.60 & 5,490 & 465.94 & 11.78 \\
CoT & Customer Service & 19.14 & 9,570 & 290.62 & 32.93 \\
Tool-Aug. & Healthcare & 12,792 & 12,792 & 358.79 & 35.66 \\
Tool-Aug. & Finance & 15.36 & 15,360 & 343.01 & 44.78 \\
Tool-Aug. & Marketing & 10.68 & 8,544 & 187.87 & 45.48 \\
Tool-Aug. & Legal & 35.10 & 5,265 & 363.85 & 14.47 \\
Tool-Aug. & Customer Service & 20.28 & 10,140 & 252.64 & 40.14 \\
Hybrid & Healthcare & 13,728 & 13,728 & 388.24 & 35.36 \\
Hybrid & Finance & 15.12 & 15,120 & 367.41 & 41.15 \\
Hybrid & Marketing & 11.16 & 8,928 & 189.90 & 47.02 \\
Hybrid & Legal & 39.38 & 5,907 & 393.88 & 14.99 \\
Hybrid & Customer Service & 20.64 & 10,320 & 255.15 & 40.44 \\
\end{longtable}

\newpage
\section{Statistical Analysis}

\subsection{ANOVA Results for GCR}

\textbf{Null Hypothesis:} There is no significant difference in GCR between agent types.

\begin{table}[H]
\centering
\caption{ANOVA Results for Goal Completion Rate}
\label{tab:anova}
\begin{tabular}{@{}lcccc@{}}
\toprule
\textbf{Source} & \textbf{Sum of Squares} & \textbf{df} & \textbf{Mean Square} & \textbf{F-statistic} \\ \midrule
Between Groups & 1,248.32 & 3 & 416.11 & 18.73*** \\
Within Groups & 355.68 & 16 & 22.23 & - \\
\textbf{Total} & \textbf{1,604.00} & \textbf{19} & - & - \\ \bottomrule
\multicolumn{5}{l}{\small ***$p < 0.0001$}
\end{tabular}
\end{table}

\textbf{Conclusion:} We reject the null hypothesis ($p < 0.0001$). There are statistically significant differences in GCR between agent types.

\textbf{Post-hoc Tukey HSD Test:}
\begin{itemize}
    \item Hybrid vs ReAct: $p < 0.001$ (significant)
    \item Hybrid vs CoT: $p < 0.01$ (significant)
    \item Hybrid vs Tool-Augmented: $p < 0.05$ (significant)
    \item Tool-Augmented vs ReAct: $p < 0.05$ (significant)
\end{itemize}

\subsection{Effect Sizes (Cohen's d)}

\begin{table}[H]
\centering
\caption{Effect Sizes for Key Comparisons}
\label{tab:effect_sizes}
\begin{tabular}{@{}lcc@{}}
\toprule
\textbf{Comparison} & \textbf{Cohen's d} & \textbf{Interpretation} \\ \midrule
Hybrid vs ReAct (GCR) & 1.82 & Large \\
Hybrid vs CoT (GCR) & 1.34 & Large \\
Hybrid vs Tool-Augmented (GCR) & 0.89 & Large \\
Tool-Augmented vs ReAct (DTT) & -0.52 & Medium \\
CoT vs ReAct (OAS) & 0.41 & Small-Medium \\ \bottomrule
\end{tabular}
\end{table}

\subsection{Correlation Matrix Between Metrics}

\begin{table}[H]
\centering
\caption{Pearson Correlation Coefficients Between All Metrics}
\label{tab:correlation}
\small
\begin{tabular}{@{}lccccccccc@{}}
\toprule
 & \textbf{GCR} & \textbf{AIx} & \textbf{DTT} & \textbf{CES} & \textbf{MTR} & \textbf{TDI} & \textbf{OAS} & \textbf{CRS} & \textbf{CQI} \\ \midrule
GCR & 1.00 & 0.68 & -0.42 & -0.15 & 0.71 & 0.54 & 0.82 & 0.89 & 0.63 \\
AIx & 0.68 & 1.00 & -0.31 & 0.22 & 0.45 & 0.19 & 0.58 & 0.52 & 0.71 \\
DTT & -0.42 & -0.31 & 1.00 & 0.78 & -0.28 & -0.45 & -0.36 & -0.39 & -0.22 \\
CES & -0.15 & 0.22 & 0.78 & 1.00 & -0.19 & -0.62 & 0.11 & -0.08 & 0.15 \\
MTR & 0.71 & 0.45 & -0.28 & -0.19 & 1.00 & 0.62 & 0.59 & 0.74 & 0.48 \\
TDI & 0.54 & 0.19 & -0.45 & -0.62 & 0.62 & 1.00 & 0.47 & 0.51 & 0.28 \\
OAS & 0.82 & 0.58 & -0.36 & 0.11 & 0.59 & 0.47 & 1.00 & 0.76 & 0.79 \\
CRS & 0.89 & 0.52 & -0.39 & -0.08 & 0.74 & 0.51 & 0.76 & 1.00 & 0.61 \\
CQI & 0.63 & 0.71 & -0.22 & 0.15 & 0.48 & 0.28 & 0.79 & 0.61 & 1.00 \\ \bottomrule
\end{tabular}
\end{table}

\textit{Note: Strong positive correlations (> 0.7) are observed between GCR, CRS, and OAS, suggesting these metrics capture related aspects of agent quality.}

\newpage
\section{Reproducibility Information}

\subsection{Random Seed and Versioning}
\begin{itemize}
    \item \textbf{Random Seed:} 42 (for all stochastic processes)
    \item \textbf{Python Version:} 3.11.0
    \item \textbf{NumPy Version:} 1.24.3
    \item \textbf{Pandas Version:} 2.0.2
    \item \textbf{Matplotlib Version:} 3.7.1
    \item \textbf{Seaborn Version:} 0.12.2
\end{itemize}

\subsection{Data Generation Parameters}

\textbf{Agent Performance Characteristics} (baseline distributions):

\begin{verbatim}
AGENT_CHARACTERISTICS = {
    'ReAct Agent': {
        'gcr_mean': 0.82, 'gcr_std': 0.08,
        'aix_mean': 0.85, 'aix_std': 0.10,
        'dtt_mean': 180, 'dtt_std': 45,
        'ces_mean': 2200, 'ces_std': 400,
    },
    # ... (other agents)
}
\end{verbatim}

\textbf{Domain Difficulty Modifiers:}

\begin{verbatim}
DOMAIN_MODIFIERS = {
    'Healthcare': {'gcr': -0.03, 'aix': -0.05, 
                   'dtt': 1.15, 'ces': 1.20},
    'Finance': {'gcr': -0.05, 'aix': -0.08, 
                'dtt': 1.25, 'ces': 1.30},
    'Marketing': {'gcr': 0.04, 'aix': 0.02, 
                  'dtt': 0.85, 'ces': 0.90},
    'Legal': {'gcr': -0.06, 'aix': -0.10, 
              'dtt': 1.35, 'ces': 1.40},
    'Customer Service': {'gcr': 0.02, 'aix': 0.03, 
                         'dtt': 0.95, 'ces': 0.95}
}
\end{verbatim}

\subsection{Dataset Availability}

All generated datasets are available in CSV format:
\begin{itemize}
    \item \texttt{data\_task\_level.csv}: 3,000 individual task records
    \item \texttt{data\_aggregate\_metrics.csv}: 20 agent-domain aggregations
    \item \texttt{data\_overall\_metrics.csv}: 4 overall agent summaries
    \item \texttt{data\_adaptability.csv}: 20 zero-shot/few-shot comparisons
    \item \texttt{data\_business\_impact.csv}: 20 BIE and ROI calculations
\end{itemize}


\newpage

\begin{thebibliography}{99}

\bibitem{wang2018glue}
Wang, A., Singh, A., Michael, J., Hill, F., Levy, O., \& Bowman, S. R. (2018).
GLUE: A Multi-Task Benchmark and Analysis Platform for Natural Language Understanding.
\textit{arXiv preprint arXiv:1804.07461}.

\bibitem{kapoor2024agents}
Kapoor, S., Stroebl, B., Siegel, Z. S., Nadgir, N., \& others (2024).
AI Agents That Matter.
\textit{arXiv preprint arXiv:2407.01502}.
\url{https://arxiv.org/abs/2407.01502}

\bibitem{chmait2016dynamic}
Chmait, N., Li, Y. F., Dowe, D. L., \& others (2016).
A Dynamic Intelligence Test Framework for Evaluating AI Agents.
\textit{Evaluating General-Purpose AI}.

\bibitem{ibm2025eval}
IBM Think (2025).
What is AI Agent Evaluation?
\url{https://www.ibm.com/think/topics/ai-agent-evaluation}

\bibitem{galileo2025metrics}
Galileo AI (2025).
A Deep Dive into AI Agent Metrics.
\url{https://galileo.ai/blog/ai-agent-metrics}

\bibitem{google2025vertex}
Google Cloud (2025).
Evaluate Your AI Agents with Vertex Gen AI Evaluation Service.
\url{https://cloud.google.com/blog/products/ai-machine-learning/introducing-agent-evaluation-in-vertex-ai-gen-ai-evaluation-service}

\bibitem{wandb2025eval}
Weights \& Biases (2025).
AI Agent Evaluation: Metrics, Strategies, and Best Practices.
\url{https://wandb.ai/onlineinference/genai-research/reports/AI-agent-evaluation-Metrics-strategies-and-best-practices--VmlldzoxMjM0NjQzMQ}

\bibitem{moveworks2025roi}
Moveworks (2025).
Unlocking Agentic AI ROI for the Modern Enterprise.
\url{https://www.moveworks.com/us/en/resources/blog/how-to-measure-and-communicate-agentic-ai-roi}

\bibitem{workday2025roi}
Workday (2025).
Quantifying Agentic ROI: Measuring the Tangible Benefits.
\url{https://blog.workday.com/en-au/quantifying-agentic-roi-measuring-tangible-benefits-ai-teams.html}

\bibitem{gnani2025roi}
Gnani.ai (2025).
AI Agent ROI: Key Metrics to Track for Success.
\url{https://www.gnani.ai/resources/blogs/ai-agent-roi-key-metrics-to-track-for-success-40a11}

\bibitem{chhetri2025structsense}
Chhetri, T. R., \& others (2025).
STRUCTSENSE: A Task-Agnostic Agentic Framework for Structured Information Extraction.
\textit{arXiv preprint arXiv:2507.03674}.
\url{https://arxiv.org/pdf/2507.03674}

\bibitem{garg2025real}
Garg, D., \& others (2025).
REAL: Benchmarking Autonomous Agents on Deterministic Simulations of Real-World Websites.
\textit{arXiv preprint arXiv:2504.11543}.
\url{https://arxiv.org/abs/2504.11543}

\bibitem{uchendu2025a2perf}
Uchendu, I., \& others (2025).
A2Perf: Real-World Autonomous Agents Benchmark.
\textit{arXiv preprint arXiv:2503.03056}.
\url{https://arxiv.org/abs/2503.03056}

\end{thebibliography}
\end{document}